\documentclass[journal]{IEEEtran}

\usepackage[utf8]{inputenc}
\usepackage{orcidlink}
\usepackage{amsmath,amssymb,amsfonts}
\usepackage{algorithmic}
\usepackage{graphicx}
\usepackage{textcomp}
\usepackage{float}
\usepackage{caption}
\usepackage{multirow}
\usepackage{mathtools}
\usepackage{bm}
\usepackage{adjustbox}
\usepackage{subfigure}
\usepackage[table]{xcolor}
\usepackage{soul}
\usepackage{gensymb}
\usepackage{authblk}
\usepackage[utf8]{inputenc}
\usepackage{booktabs}
\usepackage{nameref}
\usepackage{placeins}
\usepackage[ruled,vlined]{algorithm2e}
\SetKw{Return}{return}
\usepackage{hyperref}
\usepackage{siunitx} 
\usepackage[T1]{fontenc}
\usepackage{orcidlink}

\ifCLASSINFOpdf
\else
\fi

\begin{document}

\title{Wheel-Mounted/GNSS Fusion with AI-Aided Position Updates}

\author{Gal~Versano~\orcidlink{0009-0007-7766-2511},
        Itzik~Klein~\orcidlink{0000-0001-7846-0654}
  \thanks{G. Versano and I. Klein are with the Autonomous Navigation and Sensor Fusion Lab, Hatter Department of Marine Technologies, Charney School of Marine Sciences, University of Haifa, Israel.}}
\maketitle

\begin{abstract}
Accurate and robust localization remains a fundamental challenge for autonomous ground vehicles. In this work, we propose a hybrid neural–inertial navigation framework that integrates a wheel-mounted inertial sensors, enforced periodic trajectories, and a simple, efficient neural network capable of regressing vehicle displacement with GNSS position updates in an error-state extended Kalman filter. The periodic trajectories increase the inertial signal-to-noise ratio, allowing the network to use only inertial readings to estimate displacement. The approach is validated through real-world experiments using multiple wheel-mounted inertial sensors. Experimental results demonstrate that the proposed method achieves a significant improvement in positioning accuracy, reducing the position root mean squared error by approximately $46\%$ compared to standard wheel-mounted inertial sensor fusion with GNSS updates.
\end{abstract}


\section{Introduction}
\noindent The rapid advancement of autonomous ground vehicles (AGVs) is largely  driven by their capacity to enhance operational efficiency and enable industrial applications \cite{di2024design, tang2025path, antonyshyn2023multiple}. These platforms have become integral to specialized sectors, including precision agriculture, automated harvesting, and modern warehouse logistics. Furthermore, AGVs play a critical role in high-risk environments such as search-and-rescue operations and autonomous underground exploration \cite{kamegawa2020development, arunkumar2023review, neaz2023design}.

\noindent To obtain accurate outdoor navigation, an inertial navigation system (INS) is commonly fused with a global navigation satellite system (GNSS) receiver. INS/GNSS integration is performed using Kalman filtering techniques to estimate position, velocity, and orientation of the vehicle~\cite{groves2015principles}. The extended Kalman filter (EKF) has gained particular attention in autonomous vehicle navigation for its ability to handle nonlinear system dynamics and provide accurate trajectory estimation~\cite{housein2022extended, hsu2023hong}. Other variants of the EKF, applicable for sensor fusion, are the invariant Kalman filter (IKF) \cite{barrau2018invariant, diker2026neural, xia2024invariant}, or unscented Kalman filter (UKF)~\cite{al2021novel,meng2016covariance,levy2026adaptive}. \\
\noindent 
In parallel, deep learning approaches have been increasingly adopted in marine and aerial robotics for position estimation \cite{hou2026ins}, and hybrid methods combining neural networks with classical filtering have emerged as a promising direction, for example by using networks to denoise raw inertial data prior to EKF-based state estimation \cite{jwo2023artificial}.   In the context of GNSS/INS fusion, the potential of combining deep learning with Kalman filtering for improved navigation performance was demonstrated in~\cite{li2023exploring}. Later, \cite{hurwitz2024deep} demonstrates the benefit of fusing an INS/GNSS EKF with a neural network to enhance the overall navigation solution. More recently, a physics-informed neural network capable of regressing navigation states (position, velocity, and orientation) using only inertial sensors was proposed in~\cite{sahoo2026pidr}. These regressed navigation states were then introduced to the navigation filter as external measurements to further improve performance.\\
%
\noindent A complementary line of research focuses on mounting inertial sensors directly on vehicle wheels. In \cite{9524467} and \cite{collin2014mems}, an IMU placed at the center of a non-steering wheel is used for dead-reckoning. This configuration offers two key advantages: it can replace or complement traditional wheel encoders to mitigate error accumulation inherent to INS, and it exploits the wheel's continuous rotation to modulate and reduce the effect of sensor biases. Further, in~\cite{9508199} a comparison of three measurement models for the wheel-mounted MEMS IMU-based dead reckoning system was made. Later, simultaneous localization and terrain mapping using one wheel-mounted IMU was addressed in~\cite{9968088}. A recent wheel-mounted inertial dataset was recorded using two platforms: an omni-directional robot equipped with five IMUs, and a passenger car equipped with nine IMUs~\cite{nemec2025wheel}. This dataset was utilized in~\cite{Dusan11430518} to derive a pure inertial navigation algorithm with wheeled and chassis-mounted inertial sensors.Recently, two common practical methods for reducing inertial drift—namely, a wheel-mounted approach and enforced periodic trajectories—were combined in~\cite{versano2026wminet}. In that work, a wheel-mounted inertial neural network (WMINet) was designed to regress the mobile robot's position based solely on wheel-mounted inertial sensor readings. \\
%
%
\noindent To provide a robust and accurate solution for mobile robot navigation in scenarios where GNSS is available, we propose fusing the WMINet architecture within the navigation EKF. WMINet, leverages a wheel-mounted inertial sensing approach, while exploiting periodic motion patterns of the mobile robot.  At its core, WMINet is a simple yet effective neural network that estimates the robot’s displacement using only inertial measurements. By integrating WMINet predictions with GNSS data, we introduce an additional position update within the EKF framework.\\
Our main contribution is the introduction of a hybrid navigation approach that integrates position estimates regressed by a wheel-mounted inertial neural network (WMINet) with GNSS position measurements within an error-state extended Kalman filter. Through real-world experiments with a mobile robot with wheel-mounted IMUs, we demonstrate that the proposed approach improves the position root mean square error by approximately $46\%$ compared to a standard GNSS updates.\\
\noindent The rest of the paper is structured as follows: Section \ref{prob_for} provides a detailed explanation of the problem formulation. Section \ref{propose_app} presents our proposed approached. Section \ref{res} presents our analysis and results and Section~\ref{conc} gives the conclusions of this work.
\section{Problem Formulation}\label{prob_for}
\noindent In this section, we describe the WMINet method and present the error-state EKF, which is later implemented as part of our proposed approach.
\subsection{WMINet Framework}
\noindent
Mobile robots equipped with chassis-based inertial sensors often suffer from noisy inertial readings characterized by a low signal-to-noise ratio (SNR), particularly when moving along near-constant velocity trajectories.  To mitigate this challenge, we leverage periodic motion trajectories and utilize WMINet learning algorithm for robot positioning based solely on a wheel-mounted IMU. This periodic motion introduces continuously varying angular velocities and linear accelerations, which inherently increases the SNR of the inertial sensors. Consequently, it yields distinct, robust features that facilitate the extraction of accurate and relevant positioning information as proven in WMINet~\cite{versano2026wminet}. \\
\noindent WMINet uses a multi-head architecture combining 2D convolutional neural networks (2D-CNNs) and fully connected (FC) layers to estimate robot displacement over time. The raw IMU data is split into accelerometer and gyroscope streams, each processed independently through two convolutional layers to extract compact feature representations while preserving temporal structure.

\noindent The resulting feature maps are concatenated and passed through an additional convolutional layer, followed by two fully connected layers (512 and 32 neurons). The final output represents the estimated 2D displacement of the robot. To align with GNSS-RTK sampling, the output is structured into five one-second intervals.

\noindent Nonlinearity is introduced using the rectified linear unit (ReLU) activation function \cite{agarap2018deep}, defined as:
\begin{equation}
\text{ReLU}(\mathbf{x}) = \max(0, \mathbf{x})
\label{relufunc}
\end{equation}


\noindent A convolutional layer is defined as \cite{goodfellow2016deep}:
\begin{equation}
\mathbf{C}_{ij}^{(\ell)} = \sum_{\alpha,\beta} \boldsymbol{\omega}_{\alpha \beta}^{(\ell)} \mathbf{a}_{(i+\alpha)(j+\beta)}^{(\ell-1)} + \mathbf{b}^{(\ell)},
\label{convfunc}
\end{equation}

\noindent while a fully connected layer computes:
\begin{equation}
\mathbf{z}_{i}^{(\ell)} = \sum_{j} \boldsymbol{\omega}_{ij}^{(\ell)} \mathbf{a}_{j}^{(\ell-1)} + \mathbf{b}_{i}^{(\ell)}
\label{FC}
\end{equation}

\noindent The accelerometer (Head 1) and gyroscope (Head 2) branches are processed separately with two layers, $\mathbf{h}_{1}, \mathbf{h}_{2}$, as:
\begin{equation}
\mathbf{h}_{1} = \text{ReLU}\left(\mathbf{C}^{(0)}(\mathbf{X})\right)
\end{equation}
\begin{equation}
\mathbf{h}_{2} = \text{ReLU}\left(\mathbf{C}^{(1)}(\mathbf{h}_{1})\right)
\end{equation}

\noindent After feature extraction, both branches are concatenated:
\begin{equation}
\mathbf{h}_{\text{concat}} = \text{concat}\left(\mathbf{h}_{2}^{\text{gyro}},\ \mathbf{h}_{2}^{\text{acc}}\right)
\label{conc_func}
\end{equation}

\noindent The fused features are passed through a final convolutional layer and FC layers to produce the robot displacement:
\begin{equation}
\boldsymbol{\delta p} = \mathbf{h}_{FC}(\mathbf{C}^{(2)}(\mathbf{h}_{concat}))
\label{delta_p_eq}
\end{equation}
\noindent where $\boldsymbol{\delta p} = [\delta x, \delta y]^T$ is the estimated 2D displacement expressed in the local navigation frame.
\begin{figure*}[h]
    \centering
    \includegraphics[width=1.0\linewidth]{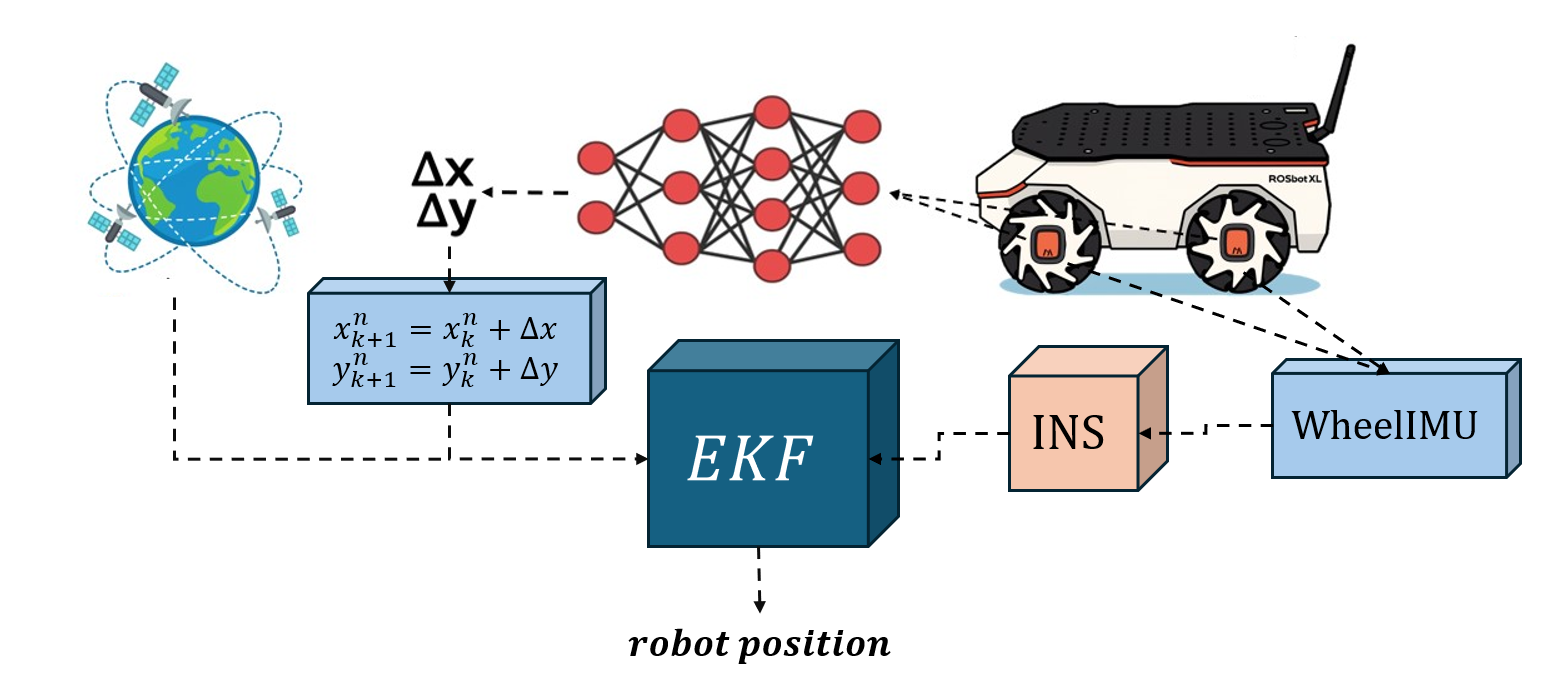}
    \caption{Block diagram of our proposed framework, illustrating the parallel inertial propagation and WMINet position update within the error-state EKF.}
    \label{fig:placeholder}
\end{figure*} 
\subsection{Navigation Filter}
\noindent Due to the nonlinear nature of the INS equations, an error-state EKF is commonly implemented in  the navigation tasks~\cite{groves2015principles}.\\

\noindent The 15-state error vector $\delta x$ is defined as:
\begin{equation} \label{eq:errorState}
\boldsymbol{\delta x} = \left[ \begin{array}{ccccc}
\boldsymbol{\delta p^{n}} & \boldsymbol{\delta v^{n}} & \boldsymbol{\phi^{n}} & \boldsymbol{b_{a}} & \boldsymbol{b_{g}} \end{array} \right]^{T} 
\end{equation}
\noindent where $\boldsymbol{\delta p^{n}}$ is the position error-states expressed in the local navigation frame, $\boldsymbol{\delta v^{n}}$ is the velocity error-states  expressed in the local navigation frame, $\boldsymbol{\phi^{n}}$ is the misalignment error-states, and $\boldsymbol{b_{a}}$ and $\boldsymbol{b_{g}}$ are the accelerometer and gyroscope bias residual errors expressed in the body frame, respectively. 

\noindent The linearized error state dynamic model is expressed using the following state-space model:
\begin{equation}
\boldsymbol{\delta \dot {x}} =\mathbf{F} \boldsymbol{\delta x} +\mathbf{G}\boldsymbol{w}, 
\end{equation}
\noindent where $\boldsymbol{F}$ is the system matrix:
\begin{equation}
\mathbf{F} = \begin{bmatrix} 
\mathbf{F}_{pp} & \mathbf{F}_{pv} & \mathbf{0}_{3 \times 3} & \mathbf{0}_{3 \times 3} & \mathbf{0}_{3 \times 3} \\
\mathbf{F}_{vp} & \mathbf{F}_{vv} & \mathbf{F}_{v\psi} & \mathbf{R}_b^n & \mathbf{0}_{3 \times 3} \\
\mathbf{F}_{\psi p} & \mathbf{F}_{\psi v} & \mathbf{F}_{\psi \psi} & \mathbf{0}_{3 \times 3} & \mathbf{R}_b^n \\
\mathbf{0}_{3 \times 3} & \mathbf{0}_{3 \times 3} & \mathbf{0}_{3 \times 3} & \mathbf{0}_{3 \times 3} & \mathbf{0}_{3 \times 3} \\
\mathbf{0}_{3 \times 3} & \mathbf{0}_{3 \times 3} & \mathbf{0}_{3 \times 3} & \mathbf{0}_{3 \times 3} & \mathbf{0}_{3 \times 3}
\end{bmatrix},
\end{equation}

\noindent $\boldsymbol{G}$ is the shaping matrix: 
\begin{equation}
\boldsymbol{\mathrm{G}} =~\left[ \begin{array}{cccc}
\boldsymbol{0_{3\times3}} & \boldsymbol{0_{3\times3}} & \boldsymbol{0_{3\times3}} & \boldsymbol{0_{3\times3}} \\ 
\mathbf{R}^{n}_{b} & \boldsymbol{0_{3\times3}} & \boldsymbol{0_{3\times3}} & \boldsymbol{0_{3\times3}} \\ 
\boldsymbol{0_{3\times3}} & \mathbf{R}^{n}_{b} & \boldsymbol{0_{3\times3}} & \boldsymbol{0_{3\times3}} \\ 
\boldsymbol{0_{3\times3}} & \boldsymbol{0_{3\times3}} & \mathbf{{I}_{3}} & \boldsymbol{0_{3\times3}} \\ 
\boldsymbol{0_{3\times3}} & \boldsymbol{0_{3\times3}} & \boldsymbol{0_{3\times3}} & \mathbf{{I}_{3}} \end{array}
\right],
\end{equation}

\noindent  $\boldsymbol{w}$ is the noise vector:
\begin{equation}
\boldsymbol{w} = \left[ \begin{array}{cccc}
\boldsymbol{w_{a}} & \boldsymbol{w_{g}} & \boldsymbol{w_{a_{b}}} & \boldsymbol{w_{g_{b}}} \end{array} \right]^{T},
\end{equation}
\noindent  $\boldsymbol{w_{a}}$ and $\boldsymbol{w_{g}}$ are the accelerometer and gyroscope measurements white noise, respectively, and $\boldsymbol{w_{a_{b}}}$ and $\boldsymbol{w_{g_{b}}}$ are the accelerometer and gyroscope biases white noise, respectively.

\noindent The Kalman filter predicts the state and its uncertainty and then updates them using external measurements, where the Kalman gain sets the relative weighting between prediction and measurement. \\
\noindent The error state covariance is propagated as~\cite{bar2011tracking}:
\begin{equation}\label{eq_p}
\mathbf{\hat{P}}^{-}_{k} = \mathbf{\Phi}_{k-1} \mathbf{\hat{P}}^{+}_{k-1} \mathbf{\Phi}_{k-1}^{T} + \mathbf{Q}_{k-1}
\end{equation}

\noindent where $\mathbf{\hat{P}}^{-}_{k}$ and $\mathbf{\hat{P}}^{+}_{k}$ are the predicted and updated covariance estimates, respectively, and $\mathbf{Q}_{k}$ is the process noise covariance matrix.

\noindent Since the INS dynamics are nonlinear, the state transition matrix $\mathbf{\Phi}$ is obtained from the linearized error-state model. In this work, it is approximated using a first-order discretization:
\begin{equation}
\mathbf{\Phi} \approx \mathbf{I} + \mathbf{F} dt
\end{equation}
where $dt$ is the sampling interval time.

\noindent The Kalman gain is computed as:
\begin{equation}
\mathbf{K}_{k} = \mathbf{\hat{P}}^{-}_{k} \mathbf{H}_{k}^{T} \left( \mathbf{H}_{k} \mathbf{\hat{P}}^{-}_{k} \mathbf{H}_{k}^{T} + \mathbf{R}_{k} \right)^{-1}
\end{equation}
where $\mathbf{H}_{k}$ is the measurement matrix and $\mathbf{R}_{k}$ is the measurement noise covariance.

\noindent The updated covariance is given by
\begin{equation}
\mathbf{\hat{P}}^{+}_{k} = \left( \mathbf{I} - \mathbf{K}_{k}\mathbf{H}_{k} \right)\mathbf{\hat{P}}^{-}_{k}
\end{equation}

\noindent Finally, the corrected error-state estimate in a closed-loop formulation is expressed as \cite{farrell2008aided}:
\begin{equation}
\boldsymbol{\delta \hat{x}}^{+}_{k} = \mathbf{K}_{k} \boldsymbol{\delta z}_{k}
\end{equation}
where $\boldsymbol{\delta z}_{k}$ is the measurement residual vector.

\section{Proposed Approach} \label{propose_app}
\begin{figure}[h]
    \centering
    \includegraphics[width=1.0\linewidth]{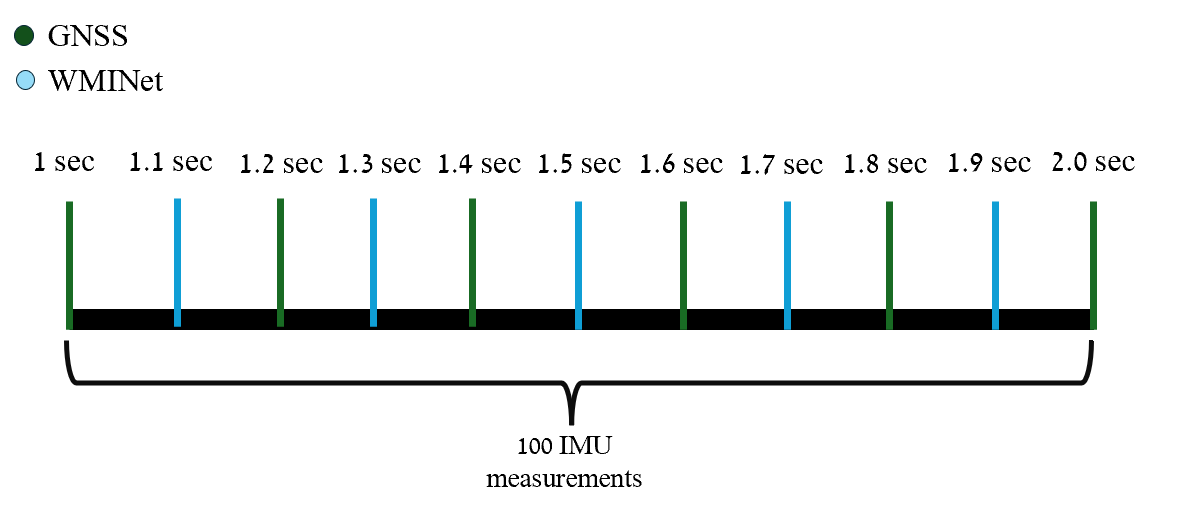}
    \caption{Update timing scheme showing the synchronization of GNSS and WMINet measurement updates at 5 Hz alongside the IMU measurements.}
    \label{time_prop}
\end{figure}

\noindent We propose a neural–inertial fusion framework in which the WMINet regressed position vector is used as an external update to the navigation filter. This fusion strategy is designed for scenarios where GNSS updates are unavailable, as well as to enhance performance when GNSS measurements are present. A block diagram of the proposed approach is shown in Figure \ref{propose_app}.
\\
In the proposed architecture, inertial measurements from the accelerometer and gyroscope are processed in two parallel branches. In the dead-reckoning branch, the IMU data is fed into the WMINet netwrok, which regresses the platform displacement over a predefined window. These estimates are then used in the dead reckoning motion model:
\begin{equation}\label{eq:2dpos21}
    \begin{aligned}
        x^n_{i} &= x^n_{i-1} + \delta x_i \\
        y^n_i &= y^n_{i-1} + \delta y_i
    \end{aligned}
\end{equation}
\noindent where $x_i$ is the $x$ coordinate position of the mobile-robot, $y_i$ is the $y$ coordinate position of the mobile robot in the navigation frame, and $i$ is the time index. \\
\noindent
Using~\eqref{eq:2dpos21}, the WMINet measurement is defined as
\begin{equation}
\mathbf{z}^{\text{WMINet}}_k = \begin{bmatrix} x_k & y_k  \end{bmatrix}^T
\end{equation}
\noindent In parallel, the same inertial measurements are used in the inertial navigation block to propagate the INS equations and the navigation filter. As a consequence, the process noise covariance (derived from the inertial sensors) is now correlated with the WMINet measurements. To account for this correlation without adding complex cross-covariance terms to the EKF pipeline, we adopt a practical engineering approach: we artificially increase the measurement noise covariance to compensate for the coupling.\\ 
\noindent
The GNSS position measurement is:
\begin{equation}
\mathbf{z}^{\text{GNSS}}_k = \begin{bmatrix} x_k & y_k  & z_k \end{bmatrix}^T
\end{equation}
To increase the overall position update rate (combined WMINet and GNSS), the WMINet updates are incorporated between two successive GNSS updates. In our dataset, both GNSS and WMINet provide measurements at 5Hz. The update timing scheme is illustrated in Figure \ref{time_prop}.
\section{Analysis and Results}\label{res}
\subsection{Dataset}
\begin{figure}[h]
\centering
    \includegraphics[width=1.0\linewidth]{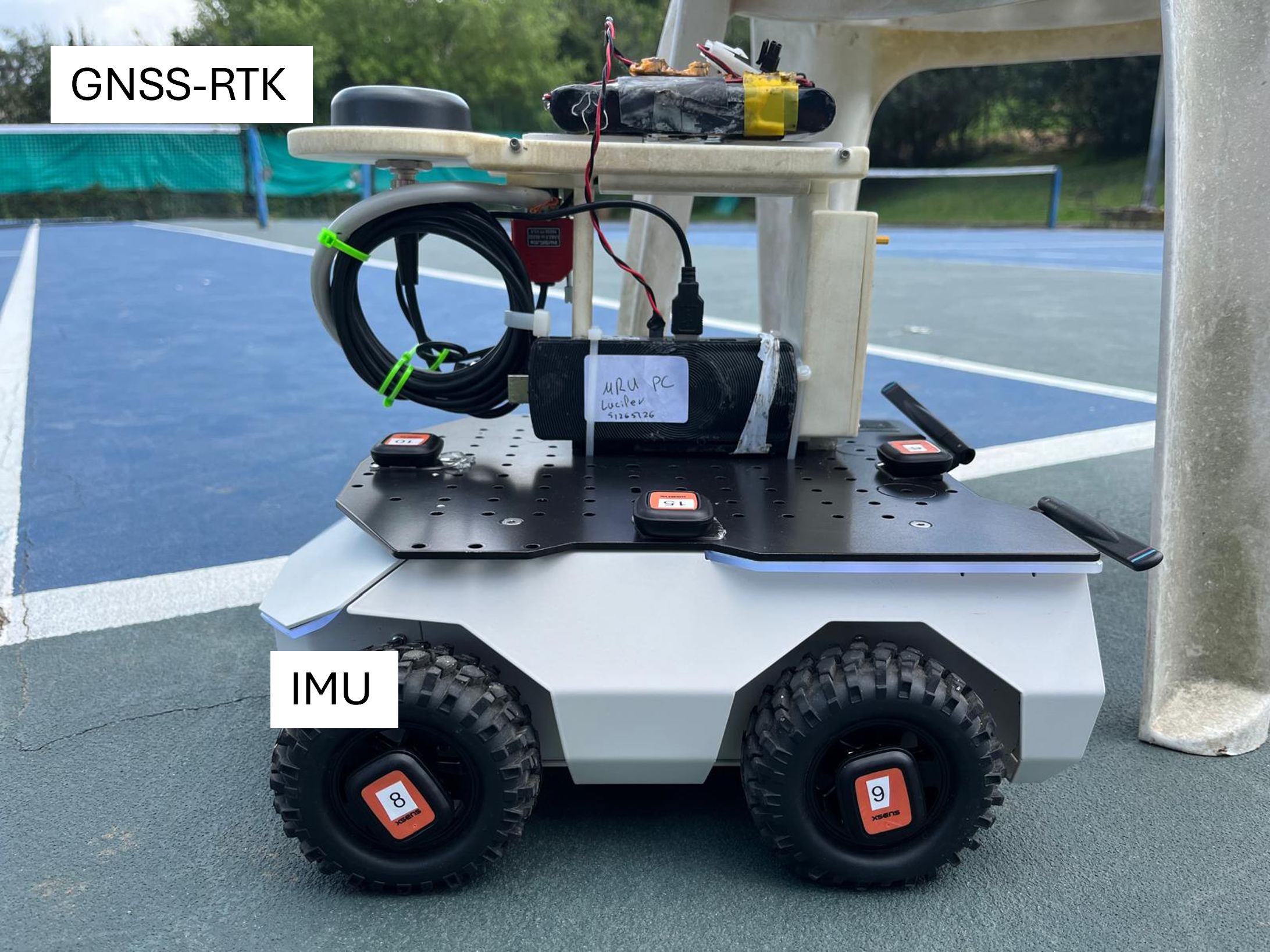}
\caption{ROSBot-XL experimental setup with wheel-mounted and chassis mounted  IMUs and a GNSS-RTK for ground-truth positioning.}
\label{setup_rosbot}
\end{figure}

\noindent 
We employ the inertial wheel-mounted dataset provided in \cite{versano2026wminet}.
A total of 26 periodic motion trajectories were collected during field experiments using ROSBot-Xl \cite{rosbotxl}, amounting to 38 minutes of data for a single IMU \cite{movella_dot} and 190 minutes for all five IMUs (four wheel-mounted units and one chassis-mounted unit). Each recording includes raw inertial measurements along with corresponding ground-truth (GT) position data captured by RTK. Figure \ref{setup_rosbot} show the setup of the experiments. We used 24 trajectories for training and 2 for testing. The training data was used to train the model, while the test data was used only to evaluate performance on unseen motion patterns. For both training and testing, we used data from two wheel-mounted IMUs as input to the neural network. Using two sensors allows us to use a distance constraint between the wheels as part of the loss function, improving the position accuracy as demonstrated in~\cite{versano2026wminet}. \\
\\
\begin{figure}[h]
    \centering
    \includegraphics[width=1.0\linewidth]{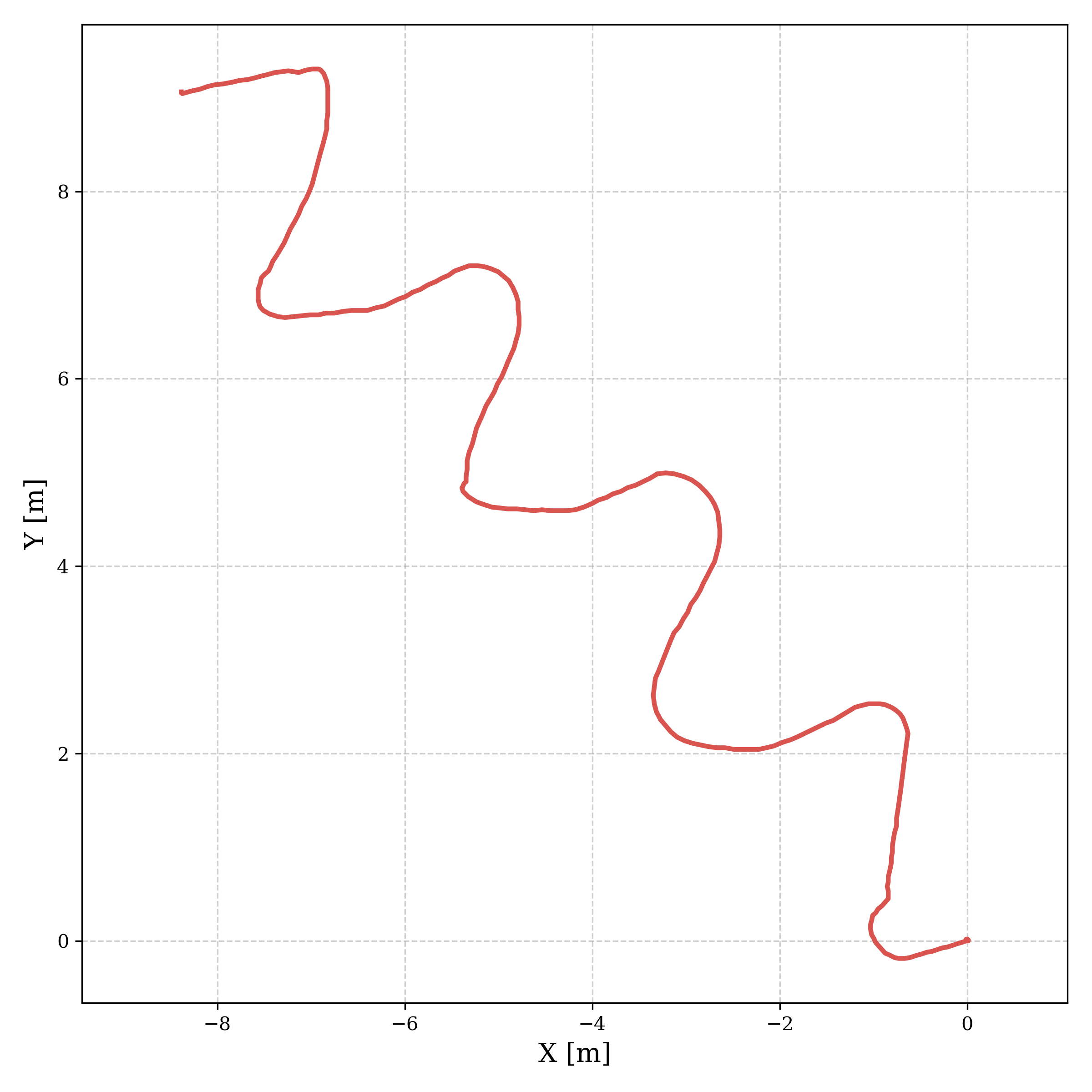}
    \caption{Horizontal position components for one of the training trajectories, demonstrating the enforced periodic motion.}
    \label{fig:placeholder}
\end{figure}
\subsection{Performance Metrics}
\noindent The evaluation metrics are chosen to quantify position estimation accuracy. The first metric is the position root mean square error (PRMSE):
\begin{equation}
\text{PRMSE} = \sqrt{\frac{1}{N}\sum_{i=1}^{N} \lVert \mathbf{x}_i - \hat{\mathbf{x}}_i \rVert^2}
\label{prmse}
\end{equation}
\noindent where $\mathbf{x}_i$ denotes the GT position vector and $\hat{\mathbf{x}}_i$ is the estimated position vector.\\
\noindent The second metric is the total distance error (TDE) defined by:
\begin{equation}
\text{TDE} (\%) = \frac{\text{PRMSE}}{D} \times 100
\label{tde}
\end{equation}
\noindent where $D$ is the total length of the  trajectory.

\subsection{Comparison with Existing Inertial Methods}

\noindent We evaluate two variants of our proposed method. as summarized below:

\begin{enumerate}
\item \textbf{WEKF}: A fusion approach combining the WMINet and EKF. There, WMINet position updates are fed into the wheel-mounted navigation EKF.
\item \textbf{EKF}: A standard EKF formulation using data from a wheel-mounted IMU.
\end{enumerate}
Note that a comparison to other inertial baselines was already presented in~\cite{versano2026wminet}, demonstrating the superiority of WMINet and therefore, it is not repeated here.
\subsection{Results}
\noindent To evaluate WEKF contribution to the  baseline EKF, we used the two test trajectories of the test dataset. Table \ref{table_PRMSE} and Table \ref{table_tde} show the PRMSE and TDE results for each of methods.

\begin{table}[h]
\centering
\caption{PRMSE for test trajectories.}
\begin{tabular}{|c|c|c|}
\hline
PRMSE [m] & WEKF (ours) &  EKF (baseline) \\ \hline
Traj.1 & 1.05    & 2.22  \\ \hline
Traj.2 & 1.25    & 2.02     \\ \hline
Average   & 1.15    & 2.12   \\ \hline
\end{tabular}
\label{table_PRMSE}
\end{table}
\noindent As can be seen from the table, the WEKF-PRMSE achieved an average result of $1.15\text{ m}$, while the EKF achieved $2.12\text{ m}$. Thus, our proposed approach improved the baseline by approximately $46\%$. This highlights the contribution of the inertial-based position updates made using WMINet.

\begin{table}[h]
\centering
\caption{ TDE for test trajectories.}
\begin{tabular}{|c|c|c|}
\hline
TDE [\%] & WEKF (ours) &  EKF (baseline) \\ \hline
 
Traj.1 & 8.75                                                               & 18.5                                                         \\ \hline
Traj.2 & 10.41                                                              & 16.83                                                        \\ \hline
Average   & 9.58                                                               & 17.65                                                        \\ \hline
\end{tabular}
\label{table_tde}
\end{table}

\begin{figure}[h]
    \centering
    \includegraphics[width=1\linewidth]{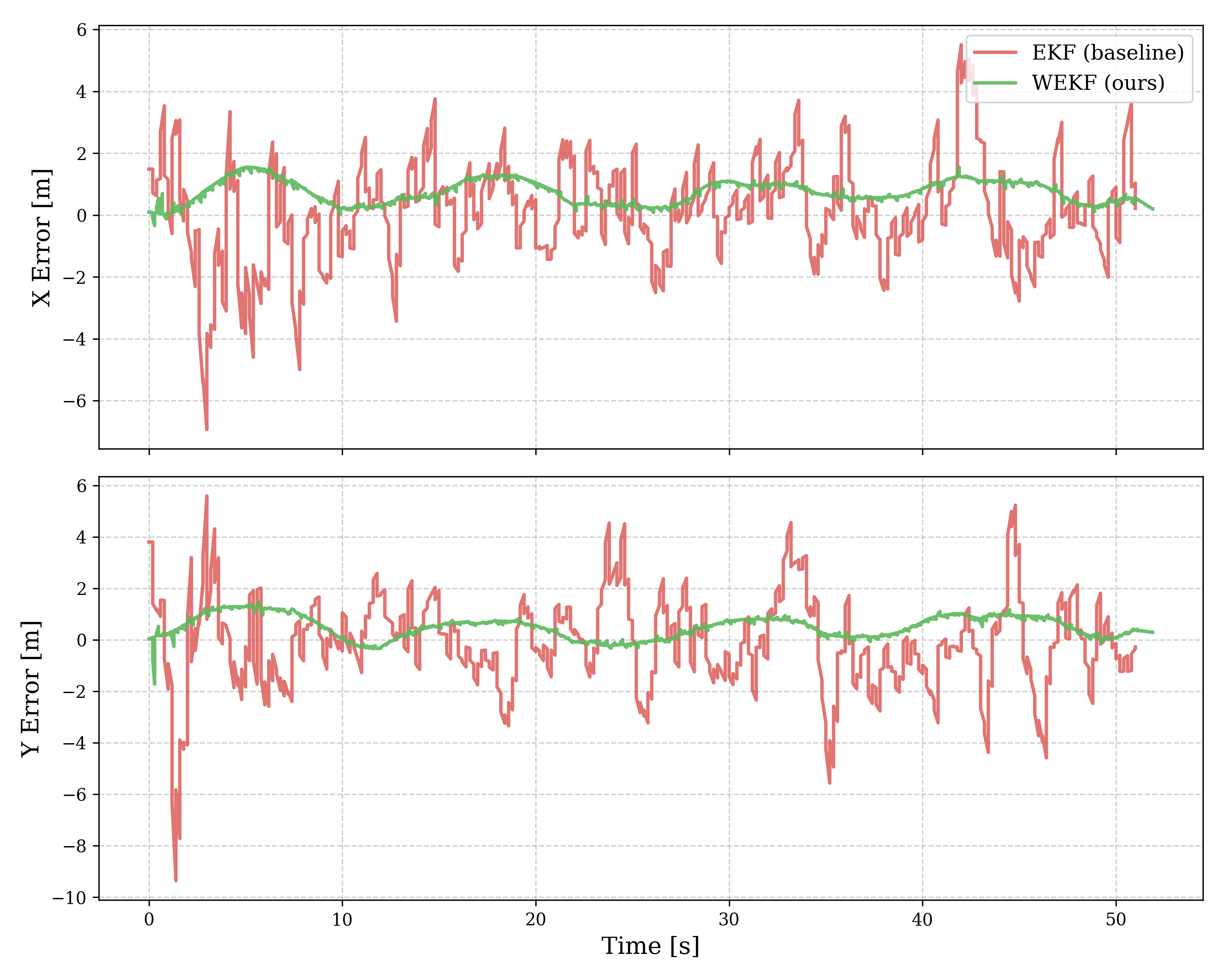}
    \caption{The time evolution of the horizontal position components for Trajectory 1.}
    \label{error_over_time}
\end{figure}

\noindent Table~\ref{table_tde} shows that the WEKF average TDE was $9.58\%$, while the baseline EKF achieved $17.65\%$, improving the total distance error by nearly a factor of two. \\
\noindent 
The time evolution of the horizontal position components for Trajectory 1 is presented in Figure \ref{error_over_time}. While the baseline EKF exhibits large fluctuations, the WEKF remains tightly bounded near zero. This demonstrates how WEKF corrections effectively smooth the estimated trajectory. \\
\noindent


\section{Conclusion}\label{conc}
\noindent In this work, we presented a hybrid neural–inertial navigation framework that uses the regressed position vector from WMINet as an additional update to the GNSS/INS navigation filter. This approach improves the localization accuracy of mobile robots equipped with wheel-mounted IMUs undergoing periodic motion. The WMINet-based position updates are introduced to the filter between two successive GNSS updates, thereby increasing the filter's update frequency.\\
\noindent Experimental results on real-world datasets demonstrate that the proposed method significantly improves positioning performance compared to a standard wheel-mounted EKF. In particular, the integration of WMINet reduces drift, stabilizes the estimation process, and achieves an average improvement of approximately $46\%$ in PRMSE. \\
\noindent Future work will focus on evaluating the framework across different mobile robot platforms and a wider variety of trajectories with varying dynamics. Additionally, we aim to address the primary limitation of this work, the restriction to periodic trajectories, by extending the methodology to handle unconstrained, arbitrary motion. \\
\noindent
Overall, this work demonstrates the benefits of combining WMINet with classical INS/GNSS navigation filter, provides a promising avenue for improving positioning accuracy of mobile robots.
\bibliographystyle{ieeetr}
\bibliography{bio}
\end{document}